\documentclass[letterpaper]{article}
\usepackage{aaai20}
\usepackage{times}
\usepackage{helvet}
\usepackage{courier}
\usepackage[hyphens]{url} 
\usepackage{graphicx}
\usepackage{graphicx}
\usepackage[draft=false]{defs-common}
\usepackage{defs-custom}

\begin{document}

\title{Conditional Level Generation and Game Blending}

\author{
 Anurag Sarkar\textsuperscript{1}, Zhihan Yang\textsuperscript{2} and Seth Cooper\textsuperscript{1} \\
 \textsuperscript{1}Northeastern University\\ 
\textsuperscript{2}Carleton College\\
sarkar.an@northeastern.edu,
yangz2@carleton.edu,
se.cooper@northeastern.edu\\
}

\maketitle

\begin{abstract}
Prior research has shown variational autoencoders (VAEs) to be useful for generating and blending game levels by learning latent representations of existing level data. We build on such models by exploring the level design affordances and applications enabled by conditional VAEs (CVAEs). CVAEs augment VAEs by allowing them to be trained using labeled data, thus enabling outputs to be generated conditioned on some input. We studied how increased control in the level generation process and the ability to produce desired outputs via training on labeled game level data could build on prior PCGML methods. Through our results of training CVAEs on levels from \textit{Super Mario Bros.}, \textit{Kid Icarus} and \textit{Mega Man}, we show that such models can assist in level design by generating levels with desired level elements and patterns as well as producing blended levels with desired combinations of games.
\end{abstract}

%================================================================================
% FIGURES

\newcolumntype{"}{@{\hskip\tabcolsep\vrule width 1pt\hskip\tabcolsep}}

\newcommand{\XFIGUREallbig}{
\begin{figure*}[h!]
\centering
\setlength\tabcolsep{1pt}
\setlength{\fboxsep}{0pt}
\setlength{\fboxrule}{1.75pt}
\begin{tabular}{ccccccccc}
\raisebox{12pt}{\rotatebox{90}{\scriptsize{SMB}}}
\includegraphics[width=0.16\columnwidth]{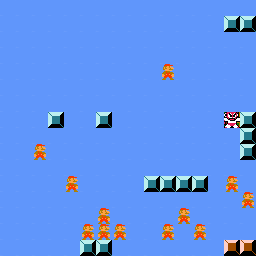} &
\includegraphics[width=0.16\columnwidth]{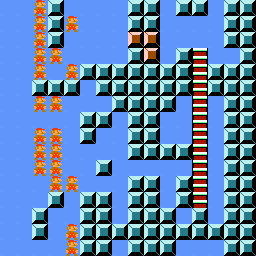} &
\includegraphics[width=0.16\columnwidth]{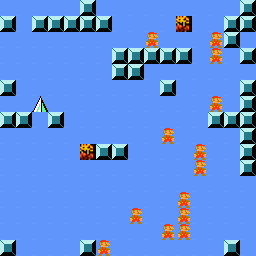} &
\includegraphics[width=0.16\columnwidth]{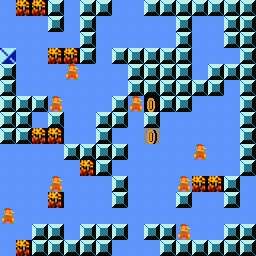} &
\fbox{\includegraphics[width=0.16\columnwidth]{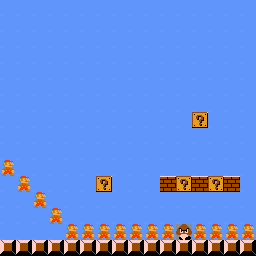}} &
\includegraphics[width=0.16\columnwidth]{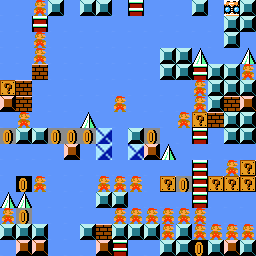} &
\includegraphics[width=0.16\columnwidth]{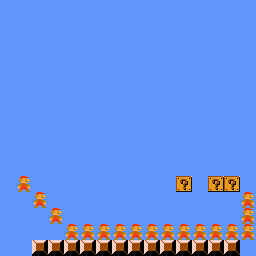} &
\includegraphics[width=0.16\columnwidth]{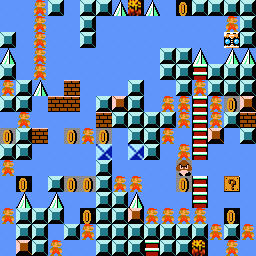} &
\\
\raisebox{15pt}{\rotatebox{90}{\scriptsize{KI}}}
\includegraphics[width=0.16\columnwidth]{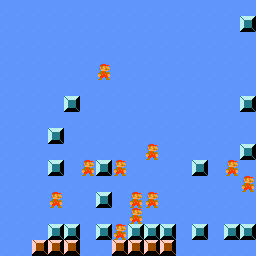} &
\includegraphics[width=0.16\columnwidth]{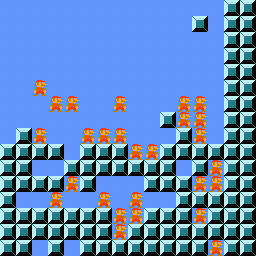} &
\fbox{\includegraphics[width=0.16\columnwidth]{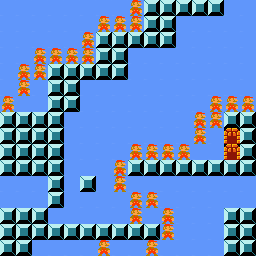}} &
\includegraphics[width=0.16\columnwidth]{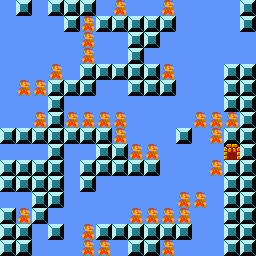} &
\includegraphics[width=0.16\columnwidth]{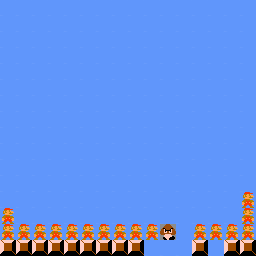} &
\includegraphics[width=0.16\columnwidth]{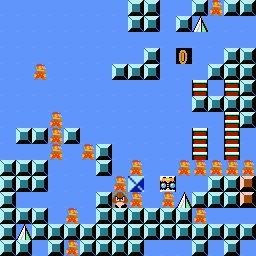} &
\includegraphics[width=0.16\columnwidth]{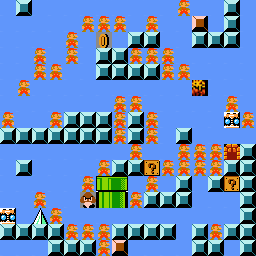} &
\includegraphics[width=0.16\columnwidth]{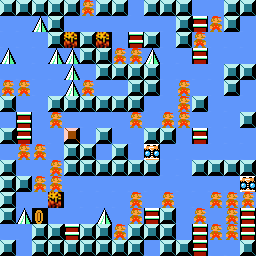} &
\\
\raisebox{15pt}{\rotatebox{90}{\scriptsize{MM}}}
\includegraphics[width=0.16\columnwidth]{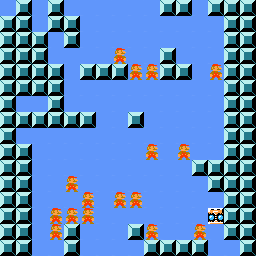} &
\fbox{\includegraphics[width=0.16\columnwidth]{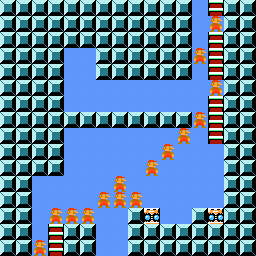}} &
\includegraphics[width=0.16\columnwidth]{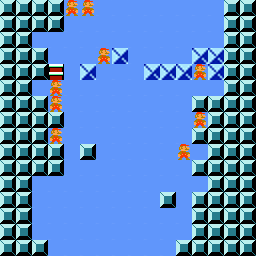} &
\includegraphics[width=0.16\columnwidth]{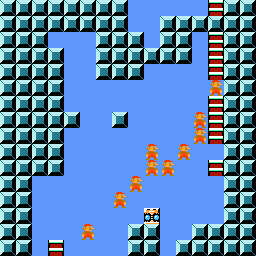} &
\includegraphics[width=0.16\columnwidth]{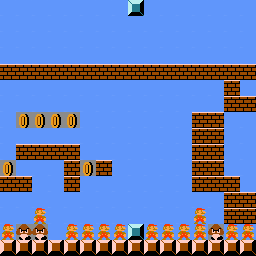} &
\includegraphics[width=0.16\columnwidth]{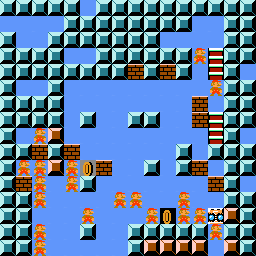} &
\includegraphics[width=0.16\columnwidth]{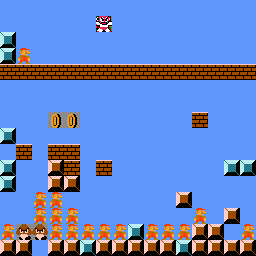} &
\includegraphics[width=0.16\columnwidth]{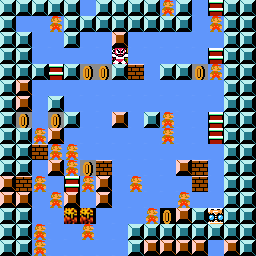} &
\\
\raisebox{10pt}{\rotatebox{90}{\scriptsize{Random}}}
\includegraphics[width=0.16\columnwidth]{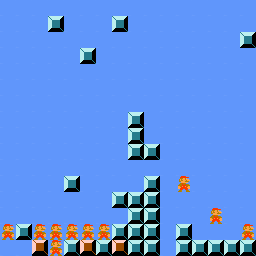} &
\includegraphics[width=0.16\columnwidth]{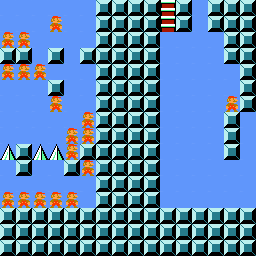} &
\includegraphics[width=0.16\columnwidth]{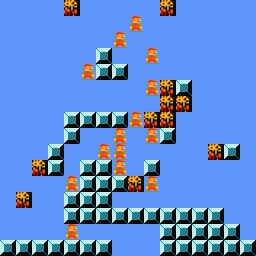} &
\includegraphics[width=0.16\columnwidth]{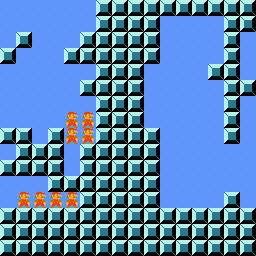} &
\includegraphics[width=0.16\columnwidth]{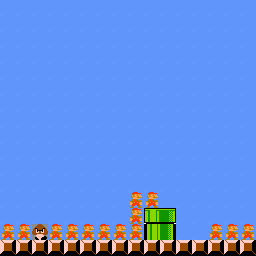} &
\includegraphics[width=0.16\columnwidth]{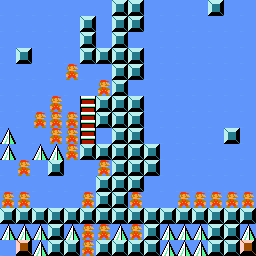} &
\includegraphics[width=0.16\columnwidth]{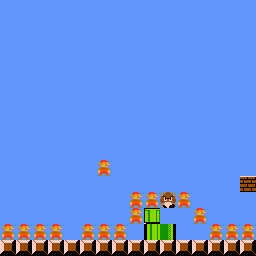} &
\includegraphics[width=0.16\columnwidth]{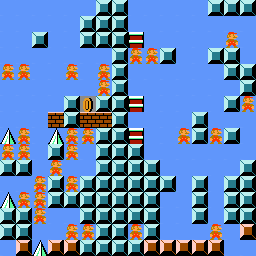} &
\\
\clabel{000} & \clabel{001} & \clabel{010} & \clabel{011} & \clabel{100} & \clabel{101} & \clabel{110} & \clabel{111}
\end{tabular}
\caption{\label{XFIGUREallbig} Segments generated by conditioning on blend labels, using an original segment from SMB (first row), KI (second) and MM (third) and a random segment (last). First, second and third elements of the label correspond to SMB, KI and MM respectively. Results were generated using the 128-dimensional model. For first 3 rows, bordered segments are originals.}
\end{figure*}
}

\newcommand{\XFIGUREsmb}{
\begin{figure}[t]
\centering
\setlength\tabcolsep{1pt}
\small
%\hspace{-1.3cm}
\begin{tabular}{c"ccccc}
\includegraphics[width=0.15\columnwidth]{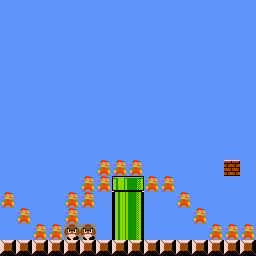} &
\includegraphics[width=0.15\columnwidth]{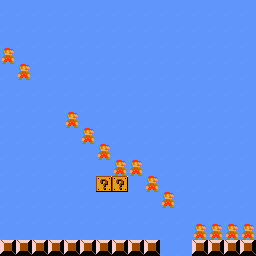} &
\includegraphics[width=0.15\columnwidth]{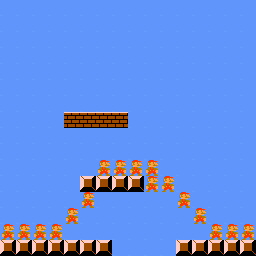} &
\includegraphics[width=0.15\columnwidth]{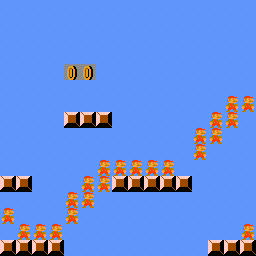} &
\includegraphics[width=0.15\columnwidth]{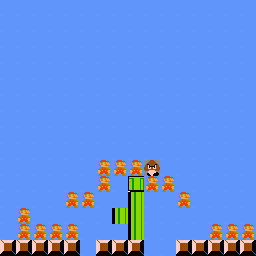} &
\includegraphics[width=0.15\columnwidth]{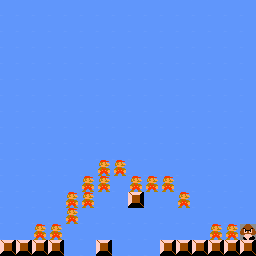}
\\
\hline
\raisebox{2em}{\small{Random}} &
\includegraphics[width=0.15\columnwidth]{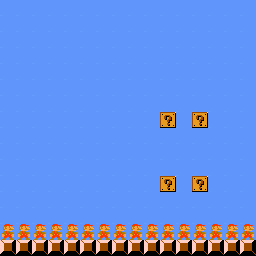} &
\includegraphics[width=0.15\columnwidth]{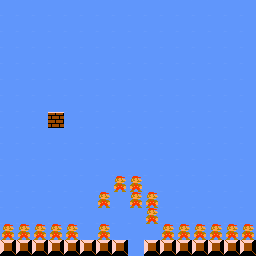} &
\includegraphics[width=0.15\columnwidth]{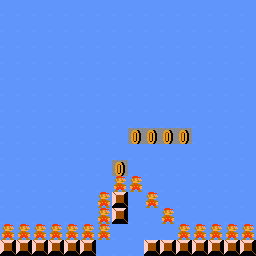} &
\includegraphics[width=0.15\columnwidth]{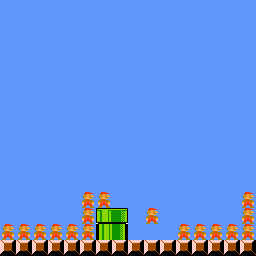} &
\includegraphics[width=0.15\columnwidth]{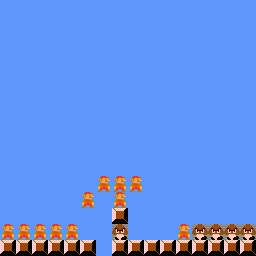}
\\
 & \clabel{000001} & \clabel{00010} & \clabel{00100} & \clabel{01000} & \clabel{10000} \\
 %& \suggestionSC{question} & \multicolumn{4}{c}{\tiny{\suggestionSC{put what the label means in this row}}}
\end{tabular}
\caption{\label{XFIGUREsmb} \hspace{-0.1cm} SMB segments generated by conditioning the original segment on the left (top) and a random vector (bottom) using the corresponding labels, as explained in Figure \ref{XFIGURElabels}.}
\end{figure}
}

\newcommand{\XFIGUREmm}{
\begin{figure}[t]
\centering
\setlength\tabcolsep{1pt}
\small
\begin{tabular}{c"ccccc}
\includegraphics[width=0.15\columnwidth]{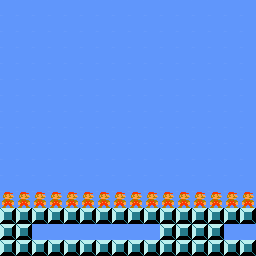} &
\includegraphics[width=0.15\columnwidth]{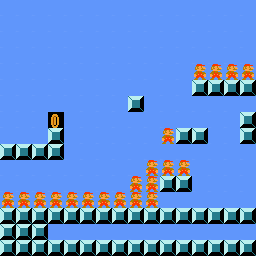} &
\includegraphics[width=0.15\columnwidth]{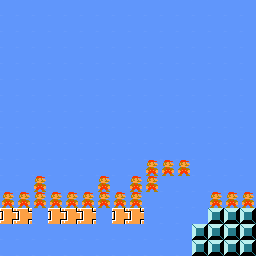} &
\includegraphics[width=0.15\columnwidth]{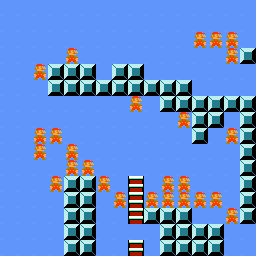} &
\includegraphics[width=0.15\columnwidth]{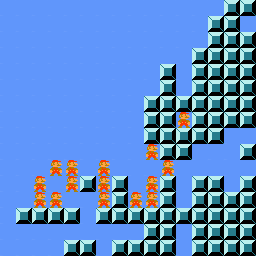} &
\includegraphics[width=0.15\columnwidth]{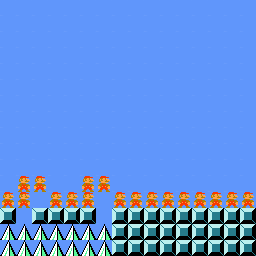}
\\
\hline
\raisebox{2em}{\small{Random}} &
\includegraphics[width=0.15\columnwidth]{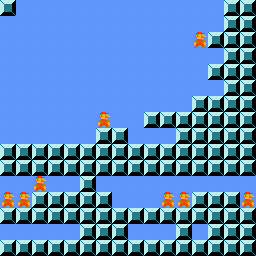} &
\includegraphics[width=0.15\columnwidth]{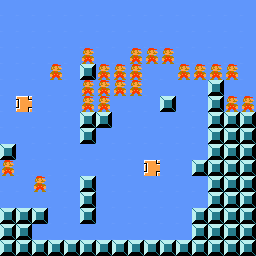} &
\includegraphics[width=0.15\columnwidth]{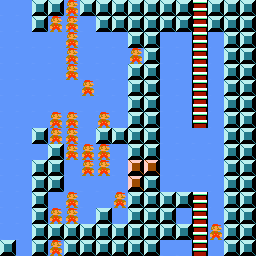} &
\includegraphics[width=0.15\columnwidth]{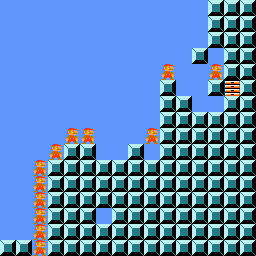} &
\includegraphics[width=0.15\columnwidth]{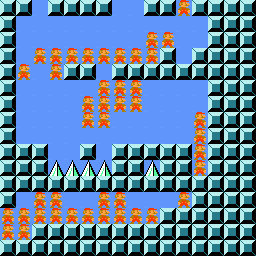}
\\
 & \clabel{00001} & \clabel{00010} & \clabel{00100} & \clabel{01000} & \clabel{10000} \\
 %& \multicolumn{5}{c}{\tiny{\suggestionSC{put what the label means in this row}}}
\end{tabular}
\caption{\label{XFIGUREmm} MM segments generated by conditioning the original segment on the left (top) and a random vector (bottom) using the corresponding labels, as explained in Figure \ref{XFIGURElabels}.}
\end{figure}
}

\newcommand{\XFIGUREki}{
\begin{figure}[t]
\centering
\setlength\tabcolsep{1pt}
\small
\setlength\tabcolsep{1pt}
\begin{tabular}{c"cccc}
\includegraphics[width=0.16\columnwidth]{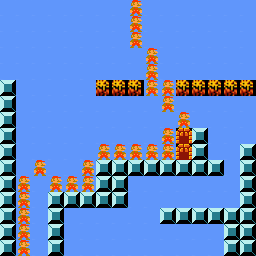} &
\includegraphics[width=0.16\columnwidth]{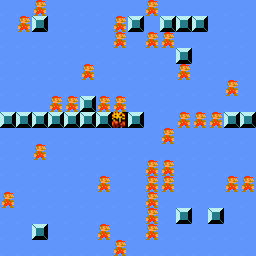} &
\includegraphics[width=0.16\columnwidth]{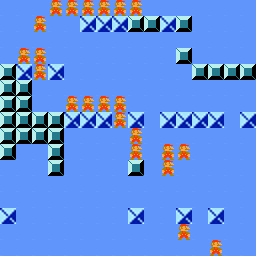} &
\includegraphics[width=0.16\columnwidth]{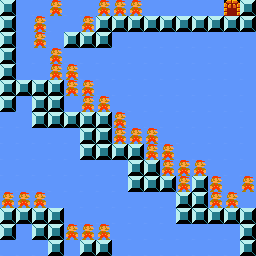} &
\includegraphics[width=0.16\columnwidth]{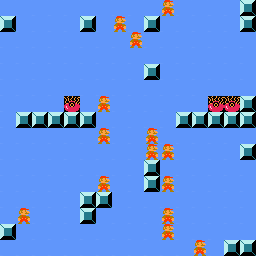}
\\
\hline
\raisebox{2em}{\small{Random}} &
\includegraphics[width=0.16\columnwidth]{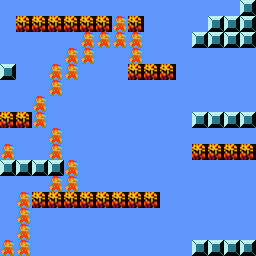} &
\includegraphics[width=0.16\columnwidth]{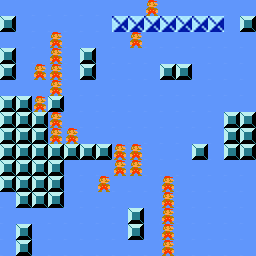} &
\includegraphics[width=0.16\columnwidth]{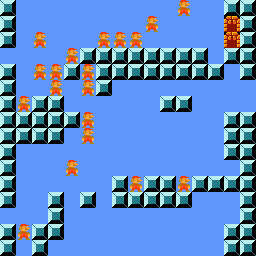} &
\includegraphics[width=0.16\columnwidth]{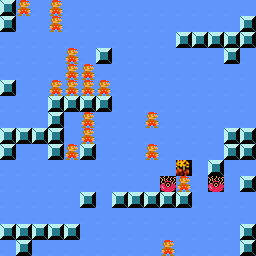}
\\
 & \clabel{0001} & \clabel{0010} & \clabel{0100} & \clabel{1000} \\
 %& \multicolumn{4}{c}{\tiny{\suggestionSC{put what the label means in this row}}}

\end{tabular}
\caption{\label{XFIGUREki} KI segments generated by conditioning the original segment on the left (top) and a random vector (bottom) using the corresponding labels as explained in Figure \ref{XFIGURElabels}.}
\end{figure}
}

\newcommand{\XFIGUREall}{
\begin{figure}[t]
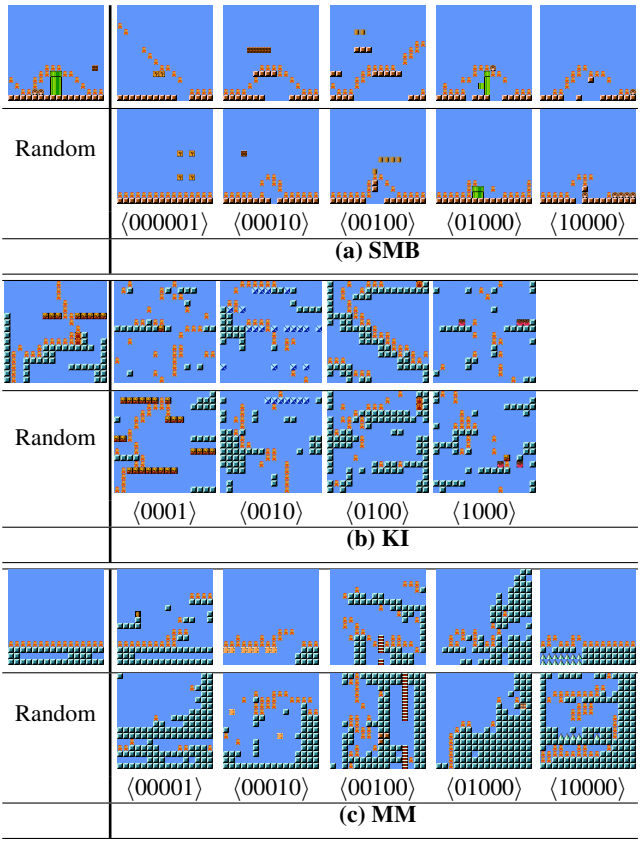

\centering
\setlength\tabcolsep{1pt}
\small
%\hspace{-1.3cm}
\begin{tabular}{c"ccccc}
\includegraphics[width=0.15\columnwidth]{figure/smb_32_orig_11010} &
\includegraphics[width=0.15\columnwidth]{figure/smb_32_00001_1} &
\includegraphics[width=0.15\columnwidth]{figure/smb_32_00010_1} &
\includegraphics[width=0.15\columnwidth]{figure/smb_32_00100_1} &
\includegraphics[width=0.15\columnwidth]{figure/smb_32_01000_1} &
\includegraphics[width=0.15\columnwidth]{figure/smb_32_10000_1}
\\
\hline
\raisebox{2em}{\small{Random}} &
\includegraphics[width=0.15\columnwidth]{figure/smbz_32_00001} &
\includegraphics[width=0.15\columnwidth]{figure/smbz_32_00010} &
\includegraphics[width=0.15\columnwidth]{figure/smbz_32_00100} &
\includegraphics[width=0.15\columnwidth]{figure/smbz_32_01000} &
\includegraphics[width=0.15\columnwidth]{figure/smbz_32_10000}
\\
 & \clabel{000001} & \clabel{00010} & \clabel{00100} & \clabel{01000} & \clabel{10000} \\
\hline
& & & \textbf{(a) SMB} & & \\[3pt]
\hline
\hline
\includegraphics[width=0.16\columnwidth]{figure/ki_32_orig_0101} &
\includegraphics[width=0.16\columnwidth]{figure/ki_32_0001_2} &
\includegraphics[width=0.16\columnwidth]{figure/ki_32_0010_2} &
\includegraphics[width=0.16\columnwidth]{figure/ki_32_0100_2} &
\includegraphics[width=0.16\columnwidth]{figure/ki_32_1000_2} &
\\
\hline
\raisebox{2em}{\small{Random}} &
\includegraphics[width=0.16\columnwidth]{figure/kiz_32_0001} &
\includegraphics[width=0.16\columnwidth]{figure/kiz_32_0010} &
\includegraphics[width=0.16\columnwidth]{figure/kiz_32_0100} &
\includegraphics[width=0.16\columnwidth]{figure/kiz_32_1000}
& \\
 & \clabel{0001} & \clabel{0010} & \clabel{0100} & \clabel{1000} & \\
\hline
& & & \textbf{(b) KI} & & \\[3pt]
\hline
\hline
\includegraphics[width=0.15\columnwidth]{figure/mm_32_orig_00000} &
\includegraphics[width=0.15\columnwidth]{figure/mm_32_00001_1} &
\includegraphics[width=0.15\columnwidth]{figure/mm_32_00010_1} &
\includegraphics[width=0.15\columnwidth]{figure/mm_32_00100_1} &
\includegraphics[width=0.15\columnwidth]{figure/mm_32_01000_1} &
\includegraphics[width=0.15\columnwidth]{figure/mm_32_10000_1}
\\
\hline
\raisebox{2em}{\small{Random}} &
\includegraphics[width=0.15\columnwidth]{figure/mmz_32_00001} &
\includegraphics[width=0.15\columnwidth]{figure/mmz_32_00010} &
\includegraphics[width=0.15\columnwidth]{figure/mmz_32_00100} &
\includegraphics[width=0.15\columnwidth]{figure/mmz_32_01000} &
\includegraphics[width=0.15\columnwidth]{figure/mmz_32_10000}
\\
 & \clabel{00001} & \clabel{00010} & \clabel{00100} & \clabel{01000} & \clabel{10000} \\
 \hline
 & & & \textbf{(c) MM} & & \\[3pt]
 \hline
\end{tabular}
\caption{\label{XFIGUREall} \hspace{-0.1cm} SMB, KI and MM segments generated by conditioning the original segment on the left (top) and a random vector (bottom) using the corresponding labels, as explained in Figure \ref{XFIGURElabels}. Changing the label changes the content generated using the same the latent vector.}
\end{figure}
}

\newcommand{\XFIGURElabels}{
\begin{figure}[t]
\centering
\begin{tabular}{ccc}
\includegraphics[width=0.16\columnwidth]{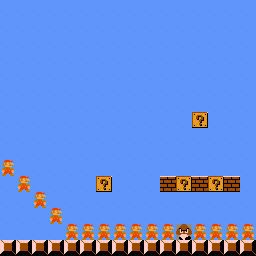} &
\includegraphics[width=0.16\columnwidth]{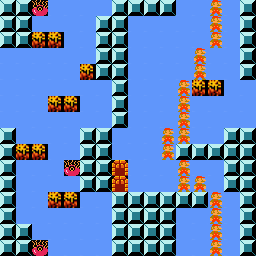} &
\includegraphics[width=0.16\columnwidth]{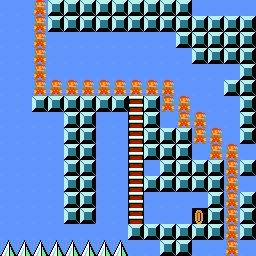} \\
SMB - \clabel{10011} & KI - \clabel{1101} & MM - \clabel{10101}
\end{tabular}
\caption{\label{XFIGURElabels} Example original segments with corresponding element labels. \textit{Super Mario Bros} (SMB)-\clabel{\textit{Enemy, Pipe, Coin, Breakable, ?-Mark}}, \textit{Kid Icarus} (KI)-\clabel{\textit{Hazard, Door, Moving Platform, Fixed Platform}}, \textit{Mega Man} (MM)-\clabel{\textit{Hazards, Door, Ladder, Platform, Collectable}}. 0/1 in labels indicate absence/presence of corresponding elements in the segment.}
\end{figure}
}

\newcommand{\XFIGUREed}{
\begin{figure}[t!]
\centering
\includegraphics[width=0.8\columnwidth]{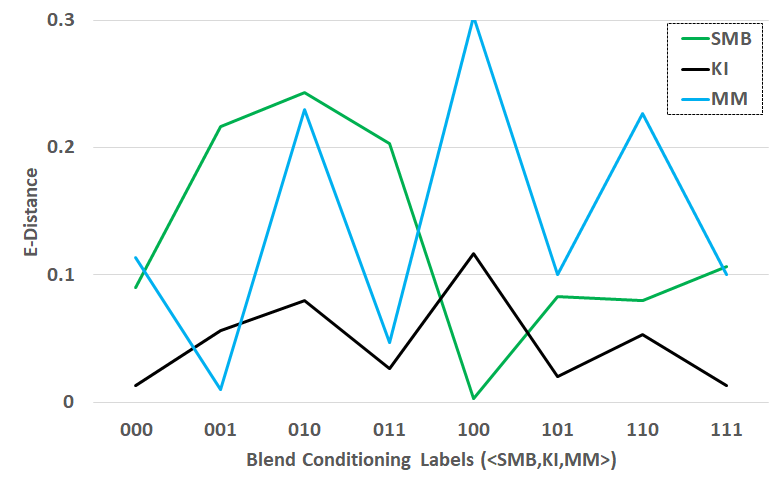}
\caption{\label{XFIGUREed} E-distances between original training distributions and distributions of 1000 levels generated using each of the blend conditioning labels.}
\end{figure}
}

\newcommand{\XFIGUREsmbplots}{
\begin{figure}[t]
\centering
\begin{tabular}{c}
\includegraphics[width=0.65\columnwidth]{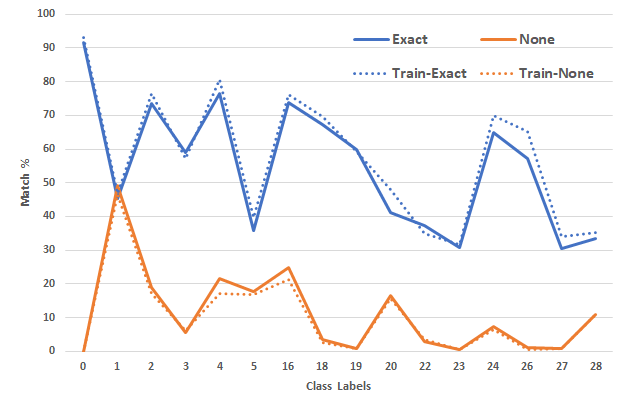} \\
\includegraphics[width=0.65\columnwidth]{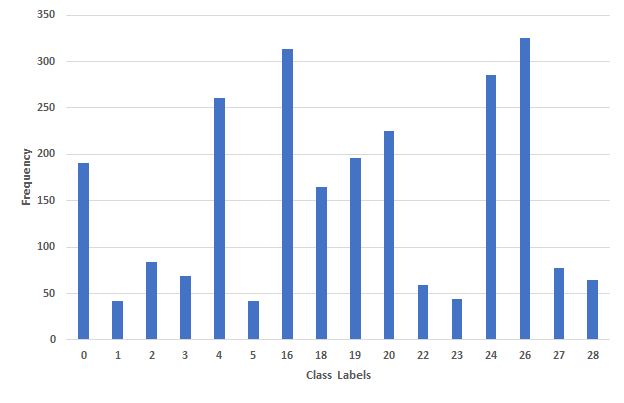} \\
\end{tabular}
\caption{\label{XFIGUREsmbplots} Results of game element conditioning for both generated and training levels of SMB (top) with frequencies of each label in the training data (bottom). X-axis values are the integer encodings of the equivalent binary label. Results shown for 16 most frequent labels in the training levels.}
\end{figure}
}

\newcommand{\XFIGUREkiplots}{
\begin{figure}[t]
\centering
\begin{tabular}{c}
\includegraphics[width=0.65\columnwidth]{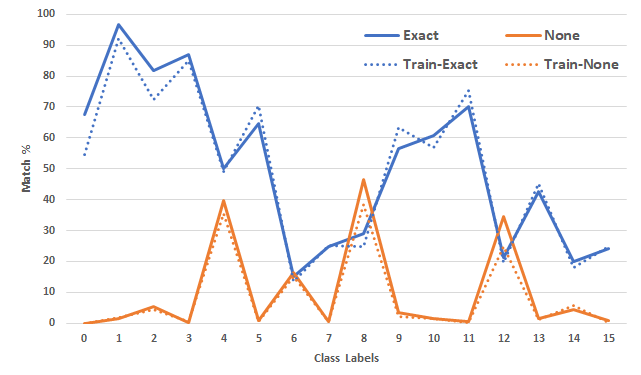} \\
\includegraphics[width=0.65\columnwidth]{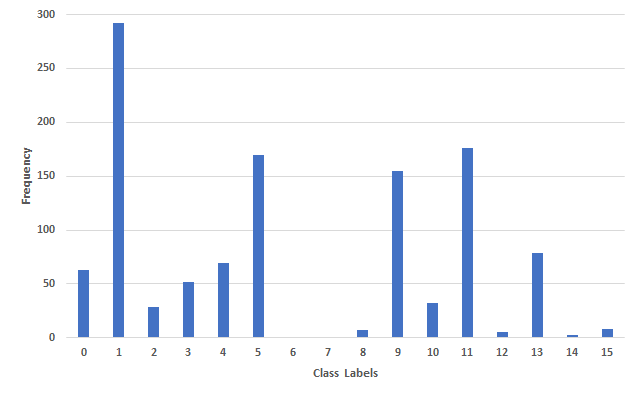} \\
\end{tabular}
\caption{\label{XFIGUREkiplots}Results of game element conditioning for both generated and training levels of KI (top) with frequencies of each label in the training data (bottom). X-axis values are the integer encodings of the equivalent binary label.}
\end{figure}
}

\newcommand{\XFIGUREmmplots}{
\begin{figure}[t]
\centering
\begin{tabular}{c}
\includegraphics[width=0.65\columnwidth]{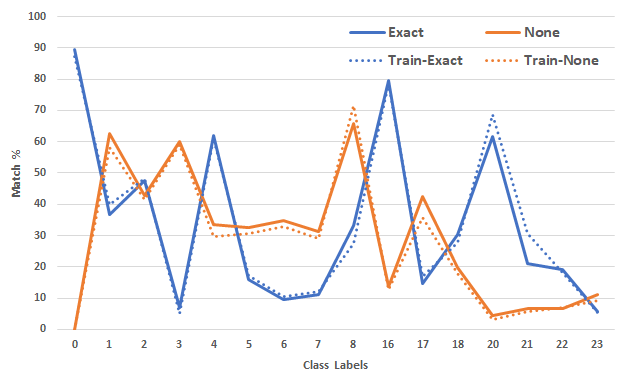} \\
\includegraphics[width=0.65\columnwidth]{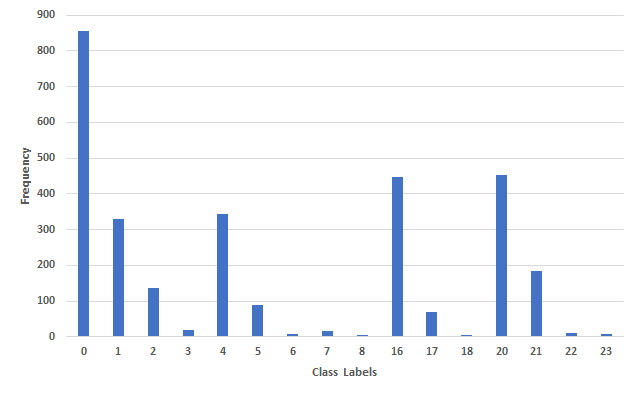} \\
\end{tabular}
\caption{\label{XFIGUREmmplots} Results of game element conditioning for both generated and training levels of MM (top) with frequencies of each label in the training data (bottom). X-axis values are the integer encodings of the equivalent binary label. Results shown for 16 most frequent labels in the training levels.}
\end{figure}
}

\newcommand{\XFIGUREsmbpats}{
\begin{figure*}[h!]
\small
\vspace{-1.2cm}
\centering
\setlength\tabcolsep{1pt}
\begin{tabular}{c"cccccccc}
\includegraphics[width=0.16\columnwidth]{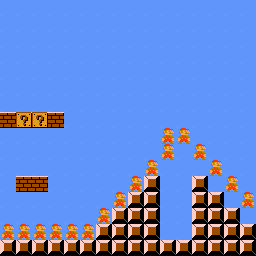} &
\includegraphics[width=0.16\columnwidth]{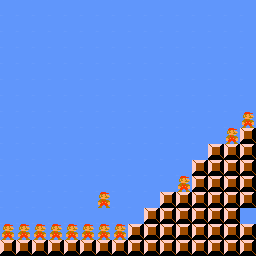} &
\includegraphics[width=0.16\columnwidth]{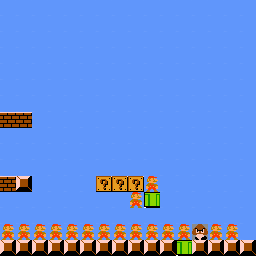}
&
\includegraphics[width=0.16\columnwidth]{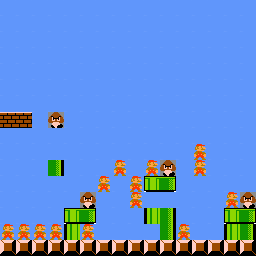} &
\includegraphics[width=0.16\columnwidth]{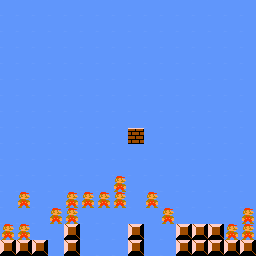} &
\includegraphics[width=0.16\columnwidth]{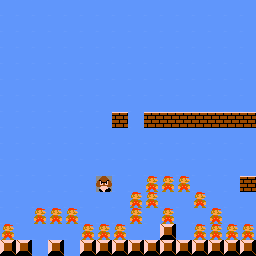} &
\includegraphics[width=0.16\columnwidth]{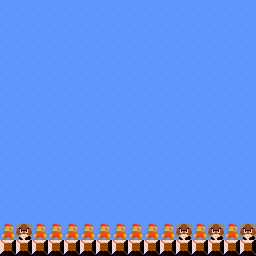} &
\includegraphics[width=0.16\columnwidth]{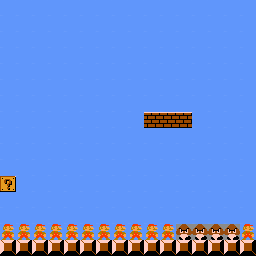} &
\includegraphics[width=0.16\columnwidth]{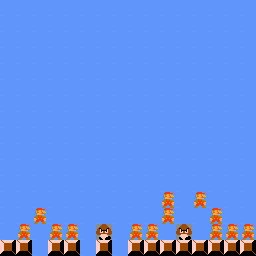}
\\
\includegraphics[width=0.16\columnwidth]{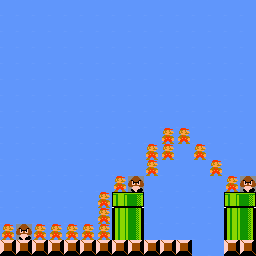} &
\includegraphics[width=0.16\columnwidth]{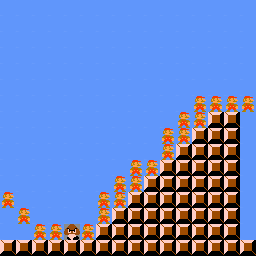} &
\includegraphics[width=0.16\columnwidth]{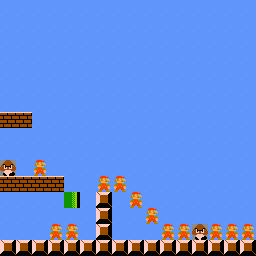} &
\includegraphics[width=0.16\columnwidth]{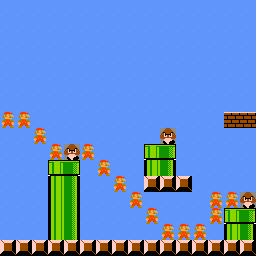} &
\includegraphics[width=0.16\columnwidth]{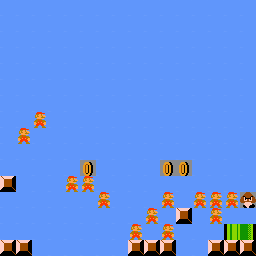} &
\includegraphics[width=0.16\columnwidth]{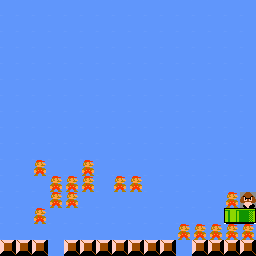} &
\includegraphics[width=0.16\columnwidth]{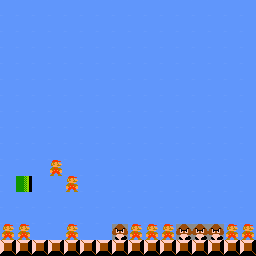} &
\includegraphics[width=0.16\columnwidth]{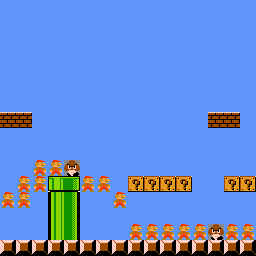} &
\includegraphics[width=0.16\columnwidth]{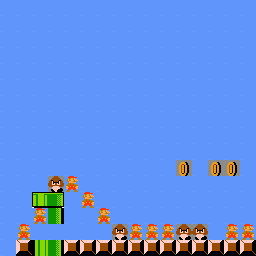}
\\
\includegraphics[width=0.16\columnwidth]{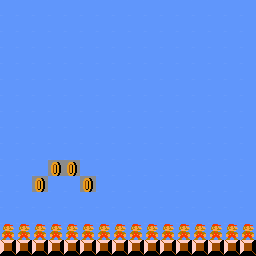} &
\includegraphics[width=0.16\columnwidth]{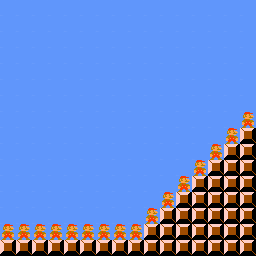} &
\includegraphics[width=0.16\columnwidth]{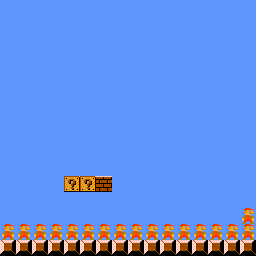} &
\includegraphics[width=0.16\columnwidth]{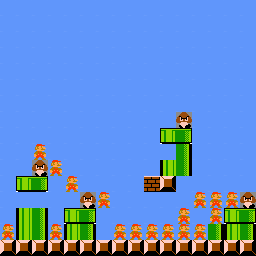} &
\includegraphics[width=0.16\columnwidth]{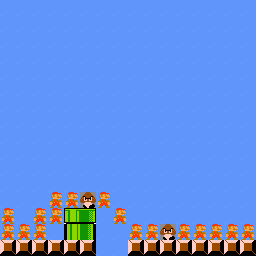} &
\includegraphics[width=0.16\columnwidth]{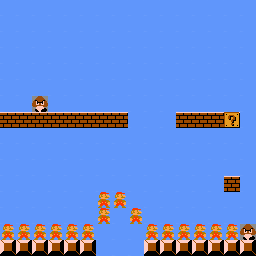} &
\includegraphics[width=0.16\columnwidth]{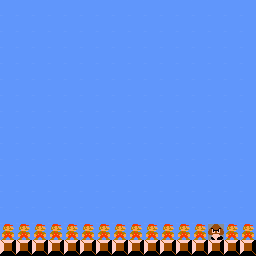} &
\includegraphics[width=0.16\columnwidth]{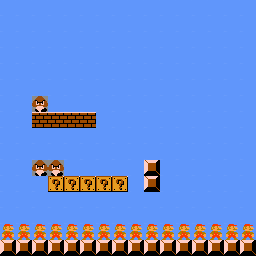} &
\includegraphics[width=0.16\columnwidth]{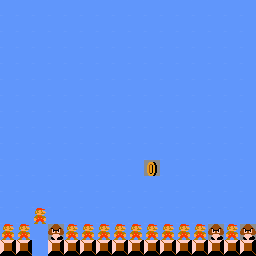}
\\
\includegraphics[width=0.16\columnwidth]{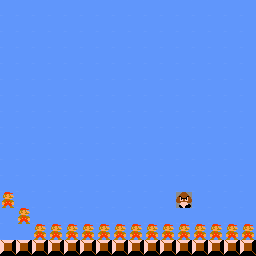} &
\includegraphics[width=0.16\columnwidth]{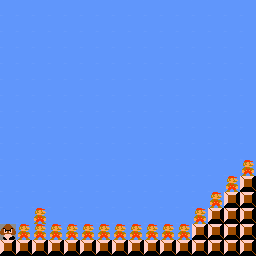} &
\includegraphics[width=0.16\columnwidth]{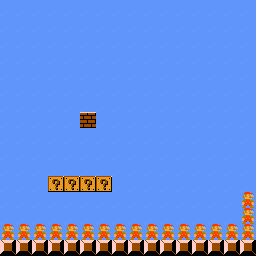} &
\includegraphics[width=0.16\columnwidth]{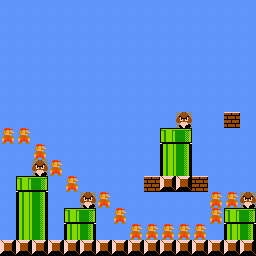} &
\includegraphics[width=0.16\columnwidth]{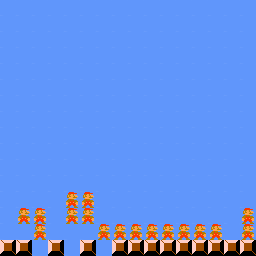} &
\includegraphics[width=0.16\columnwidth]{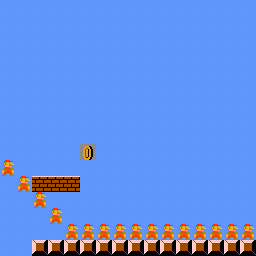} &
\includegraphics[width=0.16\columnwidth]{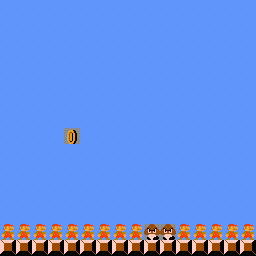} &
\includegraphics[width=0.16\columnwidth]{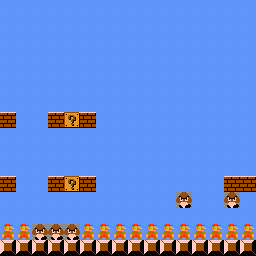} &
\includegraphics[width=0.16\columnwidth]{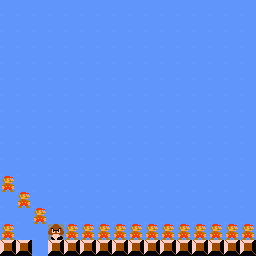}
\\
Original & \clabel{\textit{SU}} & \clabel{\textit{MP}} & \clabel{\textit{PV-NV-MP}} & \clabel{\textit{G}} & \clabel{\textit{G-MP}} & \clabel{\textit{EH}} & \clabel{\textit{EH-MP}} & \clabel{\textit{EH-G}}
\end{tabular}
\caption{\label{XFIGUREsmbpats} Example segments generated by conditioning on SMB design patterns. The first segment in each row is from the game. Every other segment in its row is generated using the same vector as the original but conditioned using the label for that column. Results shown were generated using the 32-dimensional model. Labels indicate design patterns as defined previously.}
\end{figure*}
}

%================================================================================
\newcommand{\XTABLEelems}{
\begin{table}[t]
\centering
\scriptsize
\setlength{\tabcolsep}{5pt}
\begin{tabular}{|c|c|c|c|c|c|c|}
\hline
\multicolumn{1}{|c|}{\multirow{2}{*}{}} & \multicolumn{2}{c|}{32-Dim} & \multicolumn{2}{c|}{64-Dim} & \multicolumn{2}{c|}{128-Dim} \\ \cline{2-7} 
\multicolumn{1}{|c|}{} & Exact & None & Exact & None & Exact & None \\ \hline
SMB-Rand & \textbf{33.5} & 17.6 & 32.7 & \textbf{17.1} & 27.1 & 19 \\ \hline
KI-Rand & \textbf{50.7} & \textbf{10} & 42 & 10.5 & 41.4 & 9.8 \\ \hline
MM-Rand & \textbf{17.4} & 43.2 & 15.2 & \textbf{42.5} & 14.5 & 43.7 \\ \hline
\hline
SMB-Train & \textbf{35.1} & \textbf{16.4} & 33.6 & \textbf{16.4} & 28.3 & 17.5 \\ \hline
KI-Train & \textbf{49.4} & \textbf{8.3} & 40.3 & 9.3 & 39.5 & 8.5 \\ \hline
MM-Train & \textbf{17.8} & 42.7 & 15.3 & \textbf{42.1} & 14.8 & 42.6 \\ \hline
\end{tabular}
\caption{\label{XTABLEelems} Results of conditioning randomly sampled (`Rand') and training (`Train') segments using element labels. Highest Exact values and lowest None values per game are highlighted in bold.}
\end{table}
}

\newcommand{\XTABLEelemsrand}{\begin{table}[t]
\centering
\small
\setlength{\tabcolsep}{5pt}
\begin{tabular}{|c|c|c|c|c|c|c|}
\hline
& \multicolumn{2}{c}{32-dim} & \multicolumn{2}{|c}{64-dim } & \multicolumn{2}{|c|}{128-dim} \\
\hline
 & Exact & None & Exact & None & Exact & None \\
\hline
SMB & \textbf{33.49} & 17.6 & 32.73 & \textit{17.09} & 27.05 & 18.98\\
\hline
KI & \textbf{50.74} & 9.89 & 42 & 10.46 & 41.44 & \textit{9.76}\\
\hline
MM & \textbf{17.38} & 43.24 & 15.24 & \textit{42.53} & 14.54 & 43.67\\
\hline
\end{tabular}
\caption{\label{XTABLEelemsrand} Results conditioning random vectors using element labels. Highest Exact values and lowest None values per game are highlighted in bold and italics respectively.}
\end{table}
}

\newcommand{\XTABLEelemstrain}{\begin{table}[t!]
\centering
\small
\setlength{\tabcolsep}{5pt}
\begin{tabular}{|c|c|c|c|c|c|c|}
\hline
& \multicolumn{2}{c}{32-dim} & \multicolumn{2}{|c}{64-dim } & \multicolumn{2}{|c|}{128-dim} \\
\hline
 & Exact & None & Exact & None & Exact & None \\
\hline
SMB & \textbf{35.05} & 16.43 & 33.59 & \textit{16.42} & 28.34 & 17.52\\
\hline
KI & \textbf{49.4} & \textit{8.29} & 40.32 & 9.25 & 39.47 & 8.52\\
\hline
MM & \textbf{17.75} & 42.68 & 15.27 & \textit{42.13} & 14.75 & 42.62\\
\hline
\end{tabular}
\caption{\label{XTABLEelemstrain}  Results of conditioning training levels using element labels. Highest Exact values and lowest None values per game are highlighted in bold and italics respectively.}
\end{table}
}

\newcommand{\XTABLEblend}{\begin{table}[t!]
\scriptsize
%\hspace{-0.5cm}
\centering
\setlength{\tabcolsep}{4pt}
\begin{tabular}{|c||c|c|c||c|c|c||c|c|c|}
\hline
& \multicolumn{3}{c||}{32-dim CVAE} & \multicolumn{3}{c||}{64-dim CVAE} & \multicolumn{3}{c|}{128-dim CVAE} \\
\hline
 Label & SMB & KI & MM & SMB & KI & MM & SMB & KI & MM \\
\hline
\clabel{000} &  38.7 & 18.1 & \textbf{43.2} & 31 & 20.3 & \textbf{48.7} & \textbf{41.5} & 18.2 & 20.3\\
\hline
\clabel{001} &  3.8 & 2.4 & \textbf{93.8} & 2.7 & 3.7 & \textbf{93.6} & 3.5 & 2.9 & \textbf{93.6}\\
\hline
\clabel{010} &  0.7 & \textbf{95.5} & 3.8 & 1.5 & \textbf{93.6} & 4.9 & 0.7 & \textbf{94.5} & 4.8\\
\hline
\clabel{011} &  6.8 & 22.9 & \textbf{70.3} & 7.8 & 27.5 & \textbf{64.7} & 10 & 24 & \textbf{66}\\
\hline
\clabel{100} &  \textbf{97.6} & 1.4 & 1 & \textbf{98.8} & 1.1 & 0.1 & \textbf{98.9} & 0.7 & 0.4\\
\hline
\clabel{101} & \textbf{71.9} & 2.9 & 25.2 & 20.7 & 5.2 & \textbf{74.1} & 38.1 & 2.6 & \textbf{59.3}\\
\hline
\clabel{110} &  \textbf{86.5} & 11.8 & 1.7 & \textbf{59} & 34.5 & 6.5 & \textbf{57.4} & 33.5 & 9.1\\
\hline
\clabel{111} & \textbf{56.7} & 10.3 & 33 & 32.1 & 16.8 & \textbf{51.1} & \textbf{45} & 11.1 & 43.9\\
\hline
\end{tabular}
\caption{\label{XTABLEblend} For each label, percentage of blended segments generated using that label, that was classified as the different games. Highest percentage classification for each label-dimensionality pair highlighted in bold.}
\end{table}
}
%================================================================================
\section{Introduction}
Procedural Content Generation via Machine Learning (PCGML) \cite{summerville2017procedural} has emerged as a viable means of building generative models for game levels by training on levels from existing games. While several ML approaches have been utilized for PCG such as LSTMs \cite{summerville2016mariostring}, Bayes nets \cite{guzdial2016game} and Markov models \cite{snodgrass2017learning}, a recent body of work has emerged that focuses on using latent variable models such as Generative Adversarial Networks (GANs) \cite{goodfellow2014generative} and Variational Autoencoders (VAEs) \cite{kingma2013autoencoding}. These models learn latent encodings of the input game levels, comprising a continuous latent space which can then be sampled and explored to generate new levels. Such models have been used both for level generation \cite{volz2018evolving,gutierrez2020generative} and level blending \cite{sarkar2019blending,snodgrass2020multi}. Additionally, attempts have been made to make such generation and blending controllable via latent vector evolution \cite{bontrager2018deepmasterprints}. This involves optimizing some objective function via evolutionary search in the learned latent space of the model to find vectors corresponding to levels with desired properties. Thus, controllability is achieved via evolutionary search once training has already been performed and is independent of the model. However, conditional variants of both GANs \cite{mirza2014conditional} and VAEs \cite{sohn2015learning} could enable such controllability as part of the model itself. These variants allow models to be trained on labeled data and thereby allow generation to be conditioned on input labels. Thus, when applied for PCGML, such models could use labels provided by designers to produce controllable level generators without having to define objective functions and run evolution.

Thus, we train conditional VAEs (CVAEs) on levels from \textit{Super Mario Bros.}, \textit{Kid Icarus} and \textit{Mega Man} using different sets of labels corresponding to the presence of various game elements as well as design patterns. Additionally, we train a combined model with levels labeled with the game they belong to. Our results show that CVAEs can generate levels with and without desired elements and design patterns, making it a promising model to inform future level design tools. Further, our results suggest that CVAEs can help perform controllable game blending using different labels provided during generation.

\XFIGURElabels

\section{Background}
Controllability has been the focus of much PCG research with several works developing generators that produce content based on designer preferences such as \cite{liapis_sentient_2013,alvarez2020interactive,smith2009rhythm}. Besides these systems, other works on controllable PCG have included graph-based methods \cite{valls2017graph,dormans2010adventures} and evolutionary computation \cite{togelius2010multiobjective}. Similarly, a number of PCGML works have also incorporated controllability. \citeauthoryearp{snodgrass2016controllable} used a constraint-based method to control sampling for a multi-dimensional Markov model of Mario levels while \citeauthoryearp{sarkar2018blending} used multiple LSTM models with turn-based weighting to control generation of blended Mario-Kid Icarus levels. In terms of latent variable models, 
\citeauthoryearp{volz2018evolving} used CMA-ES to evolve vectors in the latent space of a GAN trained on Mario levels. This produced levels capturing desired characteristics based on the objective used for evolution. A similar method was used by 
\citeauthoryearp{sarkar2019blending} to generate and blend levels of Mario and Kid Icarus using a VAE trained on both games. Recently,  
\citeauthoryearp{schrum2020interactive} produced a tool enabling user-controlled generation of Mario levels and Zelda-like dungeons by evolving vectors via interactive exploration of the GAN latent space. In terms of conditional models, \citeauthoryearp{torrado2019bootstrapping} combined a conditional GAN approach with self-attention mechanisms for controllable generation of GVGAI levels. Our work differs in using CVAEs instead of CGANs and not restricting to GVGAI. Moreover, their conditioning features were learned from input levels and not supplied externally as in our case, which seems more suited to future co-creative applications.

Prior work has demonstrated the utility of conceptual blending \cite{fauconnier_conceptual_1998} for generating new levels and even entire games. \citeauthoryearp{gow_towards_2015} presented a VGDL-based manual game blending framework to create new games by combining elements of existing ones. \citeauthoryearp{guzdial2016learning} similarly blended Mario levels to produce new levels. \citeauthoryearp{sarkar2018blending} and \citeauthoryearp{sarkar2019blending} used LSTMs and VAEs respectively to blend levels of Mario and Kid Icarus while \citeauthoryearp{sarkar2020exploring} extended the latter to a larger set of games and incorporated playability into blended levels. In this work, we show that game blending can also be achieved using CVAEs.

Conditional VAEs (CVAE) \cite{sohn2015learning,yan2015conditional} allow VAEs \cite{kingma2013autoencoding} to be conditioned on attributes. Unlike VAEs which are trained in an unsupervised manner, to train CVAEs, each input datapoint is associated with a label. The encoder learns to use this label to encode the input into the latent space while the decoder learns to use the label to decode the latent encoding. Thus, such models when trained on game levels could allow generation to be conditioned on designer-specified labels. Further, the same latent vector can produce different outputs by varying the conditioning labels. This enables additional affordances for level design and allows generation to be controlled without having to perform evolutionary search.

\section{Method}

\subsection{Level Data and Conditioning}
We used level data from the Video Game Level Corpus (VGLC) \cite{summerville2016vglc} for the classic NES platformers \textit{Super Mario Bros.} (SMB), \textit{Kid Icarus} (KI)  and \textit{Mega Man} (MM). For training our models, we used 16x16 level segments cropped out from the levels of each game by using a sliding window. To enable this, SMB levels and horizontal portions of MM levels were padded with 1 and 2 empty rows respectively. We finally obtained 2643 segments for SMB, 1142 for KI and 2983 for MM. Using these games, we looked at three different conditioning approaches based on 1) game elements 2) SMB design patterns and 3) game blending, motivated by wanting to generate levels containing desired elements, exhibiting desired patterns and consisting of desired combinations of games, respectively. Conditioning is accomplished by associating each input level segment with a corresponding label during the training process. In all cases, labels take the form of binary-encoded vectors. 

\subsubsection{Game Elements} 
For game elements, we used a unique set of conditioning labels for each game. The label vector length was determined by the number of different game elements considered for each game with a 0/1 value for a vector element indicating the absence/presence of the corresponding game element in the corresponding level segment. For SMB, we considered \textit{Enemy}, \textit{Pipe}, \textit{Coin}, \textit{Breakable} and \textit{Question Mark} elements, for KI, \textit{Hazard}, \textit{Door}, \textit{Moving} and \textit{Stationary Platforms} and for MM, \textit{Hazard}, \textit{Door}, \textit{Ladder}, \textit{Platforms} and \textit{Collectables}. Thus, we used 5-element binary labels for SMB and MM and 4-element labels for KI, yielding $2^5=32$ unique labels for SMB and MM and $2^4 = 16$ unique labels for KI. Example segments and their corresponding element labels are shown in Figure \ref{XFIGURElabels}.

\subsubsection{Design Patterns} For SMB design patterns, we picked 10 such patterns based on the 23 described by \citeauthoryearp{dahlskog2012patterns}. The ones we consider are:

\begin{itemize}
    \item \textit{Enemy Horde (EH):} group of 2 or more enemies
    \item \textit{Gap (G):} 1 or more gaps in the ground
    \item \textit{Pipe Valley (PV):} valley created by 2 pipes
    \item \textit{Gap Valley (GV):} valley containing a \textit{Gap}
    \item \textit{Null (empty) Valley (NV):} valley with no enemies 
    \item \textit{Enemy Valley (EV):} valley with 1 or more enemies
    \item \textit{Multi-Path (MP):} segment split into multiple parts horizontally by floating platforms
    \item \textit{Risk-Reward (RR):} segment containing a collectable guarded by an enemy
    \item \textit{Stair Up (SU):} ascending stair case pattern
    \item \textit{Stair Down (SD):} descending stair case pattern
\end{itemize}

We thus had 10-element binary labels for a total of $2^{10} = 1024$ possible unique labels though a vast majority of these do not occur in the data. For this, we only trained on SMB levels since we only used SMB design patterns.

\subsubsection{Game Blending} For game blending, we trained on segments from all 3 games taken together with labels indicating which game the segments belonged to. We used a 3-element label with the 1st, 2nd and 3rd elements indicating if the segment was from SMB, KI or MM respectively.

\subsection{Generation and Blending using CVAE}
Training CVAEs involves associating each input data instance (in our case, level segments) with a label vector. An input instance is concatenated with its label and passed through the encoder to obtain a latent vector. The same label is then concatenated with this latent vector and passed through the decoder. Thus, the encoder learns to use the provided labels to learn latent encodings of the data and similarly, the decoder learns to use the same labels to decode the encodings. Since the encoder and decoder use the labels to learn the encodings between the latent space and the data, the same latent vector could be made to produce different outputs by varying the label during generation. These affordances of the conditional VAE could inform level design and generation in two specific ways: 1) enable controllable generation by using labels to produce desired content and 2) generate variations of existing content by encoding it into latent space and decoding it using different labels. Through this work, we hoped to explore these specific affordances.

For game elements, we trained separate CVAEs for each game. In all models, both the encoder and decoder consisted of 4 fully-connected layers with ReLU activation. For SMB and MM, the conditioning labels were binary vectors of length 5 while for KI, they were of length 4. Labels for each input segment were determined by checking for the presence of the relevant game elements within that segment and assigning 0 or 1 to the corresponding element of the label. All models were trained in Pytorch \cite{paszke2017automatic} for 10000 epochs using the Adam optimizer with a learning rate of 0.001 decayed by 0.1 every 2500 epochs.

For design patterns, we trained only on SMB data. Rather than use all segments obtained by sliding the window across the levels with redundancy in terms of overlap, we only used non-overlapping segments. We found non-overlapped segments better preserve the original design patterns. Since this led to fewer segments, we additionally used VGLC level data from \textit{Super Mario Bros II: The Lost Levels}, to obtain a total of 407 segments. Labels corresponding to design patterns were assigned manually based on visual inspection. The model architecture was the same as those used for game elements, but was only trained for 5000 epochs with the learning rate decay occurring every 1250 epochs.

Finally for blending, we trained on levels from SMB, KI and MM taken together. Labels indicated the game that the levels belonged to and were of length 3 with \clabel{100}, \clabel{010} and \clabel{001} indicating SMB, KI and MM respectively. Here, we intend to achieve blending by leveraging the fact that it is possible to condition generation using labels that do not appear in the training set. For instance, though the 3 labels above are the only ones used in the training set, we could still, for example, use label \clabel{110} for conditioning and expect to generate a level that blends SMB and KI. The model architecture was similar to that used for game elements. We duplicated the number of training segments for KI to better match the number of segments for SMB and MM.

In each of the above cases, we trained 3 versions of each model, consisting of latent spaces of size 32, 64 and 128.

\XTABLEelems

\XFIGUREsmbplots

\XFIGUREkiplots

\XFIGUREmmplots

\XFIGUREall

\XFIGUREsmbpats

\XFIGUREallbig

\section{Results}
We performed a three-part evaluation focusing on each of the conditioning cases described above. Note that for KI and MM figures, we reuse certain sprites from SMB. Paths for all games are shown using a Mario character sprite.

\subsection{Game Elements}
For evaluation, for each game, we randomly sampled 1000 latent vectors and conditioned the generation for each vector using each possible label (32 for SMB and MM, 16 for KI). We then tested if the generated segments contained the elements as prescribed by the conditioning labels i.e. compared the label for a generated segment with the conditioning label used to generate it. The label for the generated segment was determined using the same method for assigning labels to training segments. We computed the percentage of segments for which the output label was an exact match as the one used for conditioning, as well as the percentage where none of the elements that the label indicated should be present were actually present in the generated segment. We also performed this evaluation for the original level segments from each game by forwarding them through the encoder and decoder using each label. Percentages of exact and none matches averaged across all labels are given in Table \ref{XTABLEelems}.

For all games and for both random samples and training levels, the 32-dimensional model leads to the highest percentage of exact matches. The 64-dimensional models do better in producing lowest percentages of no matches but not too much better than 32. Interestingly, in most cases the 128-dimensional model exhibits the worst performance for these measures. Across games, results for KI are most promising in terms of both the highest percentage of exact matches and lowest percentage of cases with no matching elements while results for MM were the worst. The average Exact match percentages seem low primarily because the model produces few exact matches when conditioning with labels not in the training data. Thus we plotted match percentages for both generated and training segments obtained when using each label. Results for the 32-dimensional model for SMB, KI and MM are given in Figures \ref{XFIGUREsmbplots}, \ref{XFIGUREkiplots} and \ref{XFIGUREmmplots} respectively. Each plot also depicts the frequency of each label in the training data. Labels are shown along the horizontal axis using the integer representation for the equivalent binary encoding. 

For all games, conditioning using labels that appear frequently in the training levels is a lot more reliable.  Example levels generated using a selection of labels for conditioning using the 32-dimensional models are shown in Figures \ref{XFIGUREall}. These show that the affordances of the CVAE enable both controllable generation and the generation of novel variations of existing content. The top rows in each of these figures demonstrate how an existing segment could be edited into a new one by simply changing the conditioning label. This holds promise for co-creative applications as a list of designer preferences regarding the elements they want in generated segments could be converted to binary conditioning labels and then used to generate the desired segments.

\subsection{Design Patterns}
Evaluating design pattern conditioning was more challenging. Unlike for game elements, where input segments could be labeled automatically by checking for elements within the segment, labels for design patterns had to be assigned manually for each segment since we lack automated methods for identifying design patterns. Consequently, we could not automatically determine if the label for a generated segment matched with the label used to generate it. Moreover, due to the low number of training segments and high number of labels, training a classifier to determine a segment's design pattern was also not feasible. Thus, we restrict our evaluation in this case to visual inspection with Figure \ref{XFIGUREsmbpats} showing example segments generated by conditioning on different labels corresponding to different combinations of design patterns. Results are shown for the 8 most common design pattern combinations in the training data. While not a robust evaluation, we see that the labels are reliable in producing the indicated design patterns in the generated segment.

\subsection{Blending}
To evaluate blending, we randomly sampled 1000 latent vectors and conditioned the generation of each using each of the 8 possible labels denoting the 8 possible combinations of the 3 games. We trained a random forest classifier on the training segments using the respective games as the class labels for the segment. We obtained a 99.12\% classification accuracy using a 80\%-20\% train-test split. Note there are 2 label types: conditioning labels appended to latents to control generation, and game labels (SMB/KI/MM) predicted by the classifier indicating which game it thinks a generated segment belongs to. For each conditioning label, we then computed the percentage of generated segments that were classified as SMB, KI or MM based on the classifier's predicted game label. When classifying segments generated via conditioning using one of the blended conditioning labels (i.e. SMB+KI, KI+MM, SMB+MM and SMB+KI+MM), we expect predicted game labels to be more spread across the 3 games versus when classifying segments conditioned using a single game conditioning label. For example, for segments generated using the SMB label (\clabel{100}), we expect a very high percentage to be classified as SMB and a very low percentage classified as others. For those generated using the blended labels, we would expect predictions with more variance. For example, for the SMB+MM label (\clabel{101}), we would expect most segments classified as SMB or MM (but not too many for either) and very few classified as KI. Results are given in Table \ref{XTABLEblend} and are in accordance with expectations. For the original game labels (i.e.\clabel{100}, \clabel{010} and \clabel{001}), a majority of segments (over 93\% in all cases) are classified as the corresponding game. This drops and is more spread out as the blended labels (i.e. those with multiple 1s) are used. Also as expected, predictions are most evenly spread out for \clabel{000} and \clabel{111}. Interestingly, in all cases other than \clabel{000}, a game is predicted 10\% or less if and only if its corresponding bit in the label is 0. These results suggest that segments generated using these labels do blend the games. Example blended levels are shown in Figure \ref{XFIGUREallbig}. We see that adding KI or MM labels to SMB segments makes them more vertical. Similarly, adding SMB labels to KI and MM gives them a more horizontal progression.

As further evaluation, for each blend label, we generated 1000 segments and computed the E-distance between them and the original training segments for each game. E-distance \cite{szekely2013energy} measures the similarity between two distributions and has been suggested as a suitable metric for comparing generative models \cite{summerville2018expanding}. The lower the E-distance between two distributions, the more similar they are. Thus, for example, we would expect the E-distance between original SMB levels and those generated using the blend conditioning label \clabel{100} to be the lowest among all labels with the value being higher for labels not containing SMB (i.e. 0 in the first label element). For computing E-distance, we used four tile-based properties: \textit{Density}, \textit{Nonlinearity}, \textit{Leniency} and \textit{Interestingness} as described in \citeauthoryearp{snodgrass2020multi}. Results for all 3 blend CVAE models are shown in Figure \ref{XFIGUREed}, averaged across the models. Results for SMB and MM are as expected with the lowest E-distance for all models being for labels \clabel{100} and \clabel{001}. For these, E-distance is higher when that game is not included in the label than when it is, as expected. Comparing to SMB, we note the sharp drop as the first label bit flips to 1 indicating inclusion of SMB in the conditioning. Similarly, comparing to MM, we note the E-distance oscillate lower to higher as the rightmost bit switches between 1 and 0, indicating inclusion of MM. Interestingly, results for KI are not as expected with label \clabel{010} producing the second highest E-distance and would be worth exploring in the future. Overall however, the E-distance trends suggest that conditioning is able to generate levels of different blends.

\XTABLEblend

\XFIGUREed

\section{Conclusion and Future Work}
We explored conditional VAEs and how their affordances can inform level design applications by allowing designers to use labels to control level generation and blending as well as edit existing levels using different labels. There are several future avenues to consider. While CVAEs enable controllability without having to run evolution, the latter could still be useful, for example, to generate content using labels that are infrequent in the training data. This could also help blending applications. Currently, the CVAE generates blended levels using given labels but the actual blended content is not controllable. Evolution could search for levels that in addition to being conditioned by a specific label, optimize a desired objective. In this work, we highlighted different CVAE affordances for controllable PCG but hope to focus on each affordance more thoroughly in the future. A user evaluation for design pattern conditioning would be useful as would playability evaluations. Additionally, this approach worked with level segments. To generate whole levels, it could be combined with the approach in \citeauthoryearp{sarkar2020sequential} that learns a sequential segment generation model where generated segments logically follow to create whole levels. Finally, we intend to incorporate such CVAE models into game design tools to enable conditional generation and design, as outlined in \citeauthoryearp{sarkar2020towards}.

%In generating diverse content from a single latent vector by changing labels, CVAEs have much in common with Quality-Diversity evolutionary approaches \cite{pugh2016quality}. Though unlike QD, the CVAE does not explicitly partition the latent space, the notion of different labels corresponding to different types of content can be viewed as analogous to different niches in QD search spaces characterizing different behaviors. In the future, we would thus like to explore how these methods may be combined to inform novel PCGML approaches. 
%\todo{controllability via disentanglement}

%\clearpage

\bibliographystyle{aaai}
\bibliography{refs-custom}

\end{document}